\documentclass[10pt,twocolumn,letterpaper]{article}

\usepackage{cvpr}
\usepackage{times}
\usepackage{epsfig}
\usepackage{graphicx}
\usepackage{amsmath}
\usepackage{amssymb}
\usepackage{url}
\usepackage{times}
\usepackage{graphicx}
\usepackage{cite}
\usepackage[ruled,lined,slide]{algorithm2e}
\usepackage{algpseudocode}
\usepackage{enumitem}
\usepackage{subfigure}
\usepackage{epstopdf}
\usepackage{lipsum}
\usepackage{multirow}
\usepackage{dsfont}
\usepackage[toc,page]{appendix}

\usepackage[pagebackref=true,breaklinks=true,letterpaper=true,colorlinks,bookmarks=false]{hyperref}

\cvprfinalcopy 

\def\cvprPaperID{0913} 

\ifcvprfinal\fi
\begin{document}

\title{Refining Architectures of Deep Convolutional Neural Networks}

\author{\fontsize{12pt}{12pt}\selectfont Sukrit Shankar$^*$ $~$ Duncan Robertson$^\dag$ $~$ Yani Ioannou$^*$ $~$ Antonio Criminisi$^\dag$ $~$ Roberto Cipolla$^*$  \\
$^*$ \fontsize{11pt}{11pt}\selectfont  Machine Intelligence Lab, University of Cambridge, UK \\
$^\dag$ \fontsize{11pt}{11pt}\selectfont  Microsoft Research Cambridge, UK \\
{\tt\footnotesize ss965@cam.ac.uk | a-durobe@microsoft.com   |  yai20@cam.ac.uk  |  antcrim@microsoft.com   |  rc10001@cam.ac.uk }
}

\maketitle

\begin{abstract}
\fontsize{10pt}{12pt}\selectfont
{
Deep Convolutional Neural Networks (CNNs) have recently evinced immense success for various image recognition tasks \cite{krizhevsky2012imagenet, zhou2014learning}. However, a question of paramount importance is somewhat unanswered in deep learning research - is the selected CNN optimal for the dataset in terms of accuracy and model size?

   
$~~$ In this paper, we intend to answer this question and introduce a novel strategy that alters the architecture of a given CNN for a specified dataset, to potentially enhance the original accuracy while possibly reducing the model size. We use two operations for architecture refinement, viz. stretching and symmetrical splitting. Stretching increases the number of hidden units (nodes) in a  given CNN layer, while a symmetrical split of say $K$ between two layers separates the input and output channels into $K$ equal groups, and connects only the corresponding input-output channel groups.  Our procedure starts with a pre-trained CNN for a given dataset, and optimally decides the stretch and split factors across the network to refine the architecture. We empirically demonstrate the necessity of the two operations. 

$~~$ We evaluate our approach on two natural scenes attributes datasets, SUN Attributes \cite{patterson2012sun} and CAMIT-NSAD \cite{Shankar_2015_CVPR}, with architectures of GoogleNet and VGG-11, that are quite contrasting in their construction. We justify our choice of datasets, and show that they are interestingly distinct from each other, and together pose a challenge to our architectural refinement algorithm. Our results substantiate the usefulness of the proposed method. 

}

\end{abstract}

\section{Introduction}  \label{sec_intro}
Deep Convolutional Neural Networks (CNNs) have recently shown immense success for various image recognition tasks, such as object recognition \cite{krizhevsky2012imagenet, szegedy2014going}, recognition of man-made places \cite{zhou2014learning}, prediction of natural scenes attributes \cite{Shankar_2015_CVPR} and discerning of facial attributes \cite{liu2014deep}. Out of the many CNN architectures, AlexNet \cite{krizhevsky2012imagenet}, GoogleNet \cite{szegedy2014going} and VGG \cite{simonyan2014very} can be considered as the most popular ones, based on their impressive performance across a variety of datasets. While architectures of AlexNet and GoogleNet have been carefully designed for the large-scale ImageNet \cite{deng2009imagenet} dataset, VGG can be seen as being relatively more generic in curation. Irrespective of whether an architecture has been hand-curated for a given dataset or not, all of them show significant redundancy in their parameter space \cite{liu2015sparse, jaderberg2014speeding}, i.e. for a significantly reduced number of parameters (sometimes as high as $90\%$ reduction), a given architecture might achieve nearly the same accuracy as obtained with the entire set of parameters. Researchers have utilized this fact to speed up inference by estimating the set of parameters that can be zeroed out with a minimal loss in original accuracy. This is typically done through sparse optimization techniques \cite{liu2015sparse} and low-rank procedures \cite{jaderberg2014speeding}. 

\textbf{Envisaging a real-world scenario:} We try to envisage a real-world scenario. A user has a new sizeable image dataset, which he wants to train with a CNN. He would typically try out famous CNN architectures like AlexNet, GoogleNet, VGG-11, VGG-16, VGG-19 and then select the one which gives maximum accuracy. In case his application prioritizes a reduced model size (number of parameters) as compared to the accuracy (e.g. applications for mobile platforms and embedded systems), he will try to strike a manual trade-off of how much he wants to sacrifice the accuracy for a reduction in model size. 

\textit{Once he makes his choice, what if his finally selected architecture could be altered so as to potentially give a better accuracy while also reducing the model size ?} In this paper, we aim to target such a scenario. Note that in all cases,  the user can further choose to apply one of the sparsification techniques such as  \cite{liu2015sparse} to significantly reduce the model size with a slight decrease in accuracy. We now formally define our \textbf{problem statement} as follows :

\textit{Given a pre-trained CNN for a specific dataset, refine the architecture in order to potentially increase the accuracy while possibly reducing the model size. }

\begin{figure}[!htb]
\begin{center}
   \includegraphics[width=0.95\linewidth, height=0.14\textheight]{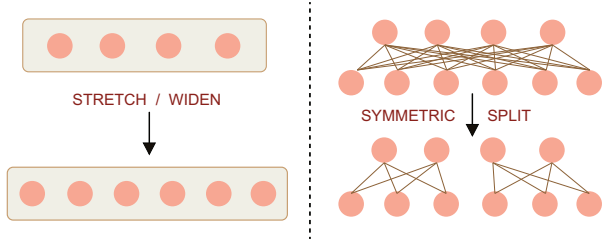}
\end{center}
   \caption{\fontsize{8.5pt}{10pt}\selectfont  \textbf{Operations considered for our approach:  } We consider two operations, viz. stretch (\textit{left}) and symmetric split (\textit{right}), for architectural refinement of a CNN. Stretching refers to increase in number of hidden units (nodes) for a  given layer, without changing its connection pattern to the previous or the next layer. A stretch by a factor of 1.5 is shown here. A symmetrical split of say $K$ between two layers separates the input and output channels into $K$ equal groups, and the corresponding input and output channel groups are connected. A symmetric split of $2$ is shown here. Symmetrical split is implemented as the \textit{group} parameter in Caffe \cite{jia2014caffe}. }
\label{fig_refinement_operations} 
\end{figure}

\textbf{Operations for CNN architecture refinement: } One may now ask what is exactly meant by the \textit{refinement of a CNN architecture}. On a broader level, refining a CNN architecture can involve altering one or more of  the following: the number of hidden units (nodes) in any layer, the connection pattern between any two layers, and the depth of the network. On a relatively  finer  level, one might think of changing the  convolution kernel size, pooling strategies and stride values to refine an architecture.
 
In this paper, we consider the task of CNN architecture refinement on a broader level. Since we embark on such a problem in this work, we only consider two operations, viz. stretch and symmetric split. Stretching refers to increase in number of hidden units (nodes) for a  given layer, while a symmetrical split of say $K$ between two layers separates the input and output channels into $K$ equal groups, and the $k^{th}$ input channel group is only connected to the $k^{th}$ output channel group\footnote{For better understanding, we give an example of symmetric splitting with convolutional layers. Let a convolutional layer $conv_1$ having $96$ outputs  be connected to $conv_2$ having $256$ outputs.  Then there are $96 \times 256$ input connections for $conv_2$, each connection having a filter of square size ($11 \times 11 ~~ say$). A splitting of $2$ for $conv_2$ divides input connections of $conv_2$ into $2$ symmetric groups, such that the first / second $48$ outputs of $conv_1$ only get connected to the first / second $128$ outputs of $conv_2$. }.  Please see Fig ~\ref{fig_refinement_operations} for an illustration of these operations.  We do \textit{not} consider the other plausible operations for architectural refinement of CNN; for instance, arbitrary connection patterns between two layers (instead of just symmetric splitting), reducing the number of nodes in a layer, and alteration in the depth of the network. 

\textbf{Intuition behind our approach: } The main idea behind our approach is to best separate the classes of a dataset, assuming a constant depth of the network. Our method starts with a pre-trained CNN, and studies separation between classes at each convolutional layer. Based on the nature of the dataset, separation between some classes may be more at lower layers, while for others, may be lesser at lower layers. Similar variations may be seen at deeper layers. In comparison to its previous layer, a given layer can increase the class separation for some class pairs, while decreasing for others. The number of class pairs for which the class separation increases contributes to the \textit{stretching / widening} of the layer; while the number of class pairs where the class separation decreases contributes to the \textit{symmetric splitting} of the layer inputs. Thus, both stretch and split operations can be simultaneously applied to each layer.  The amount of stretch or split is not only decided by how the layer affects the class separation, but also by the class separation capacity of the subsequent layers. Once the stretch and split factors are estimated, they are applied to original CNN for architectural refinement. The refined architecture is then trained again (from scratch) on the same dataset. Section ~\ref{sec_approach} provides complete details of our proposed approach. 


\textbf{Our contribution(s): } Our major contributions can now be summarized as follows : 
\begin{enumerate}
\fontsize{9.5pt}{11pt}\selectfont
{
\item For a given pre-trained CNN, we introduce the problem of refining network architecture so as to potentially enhance the accuracy while possibly reducing the required number of parameters. 
\item We introduce  a strategy that starts with a pre-trained CNN, and uses stretch and symmetric split (Fig ~\ref{fig_refinement_operations}) operations for CNN architecture refinement. 
}
\end{enumerate}

\section{Related Work} \label{sec_related_work}
 Deep Convolutional Neural Networks (CNNs) have experienced a recent surge in computer vision research due to their immense success for visual recognition tasks \cite{krizhevsky2012imagenet, zhou2014learning}. Given a sizeable training set, CNNs have proven to be far more robust as compared to the hand-crafted low-level features like Histogram of Oriented Gradients (HOG) \cite{dalal2005histograms}, color histograms, gist descriptors \cite{oliva2006building} and the like. For visual recognition, CNNs provide impressive performance for recognition of objects \cite{szegedy2014going}, man-made places \cite{zhou2014learning}, attributes of natural scenes \cite{Shankar_2015_CVPR} and facial attributes \cite{liu2014deep}. However, learning an optimal CNN architecture for a given dataset is largely an open problem. Moreover, it is less known, how to find if the selected CNN is optimal for the dataset in terms of accuracy and model size or not.
 
\begin{figure}[!htb]
\begin{center}
  \includegraphics[width=0.95\linewidth, height=0.14\textheight]{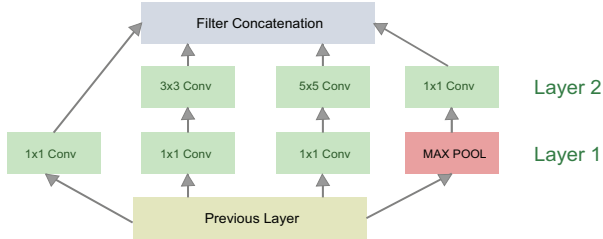}
\end{center}
   \caption{\fontsize{8.5pt}{10pt}\selectfont  \textbf{Inception Module of GoogleNet \cite{szegedy2014going} : } The inception module is an intrinsic component of the GoogleNet architecture. GoogleNet has 9 inception modules named as 3a, 3b, 4a, 4b, 4c, 4d, 4e, 5a, 5b connected one after another. The inception module has two layers and 6 convolutional blocks (green blocks), connected as shown in the figure.  As an implementation perspective of our approach with GoogleNet, for a convolutional block $l$ in Layer 1, the subsequent blocks are all convolutional blocks in layer 2, irrespective of the connection pattern. This is done for ease in the computation of \eqref{eqn_1} and \eqref{eqn_2}. However,  for a given convolutional block $l$ in a layer of inception module, its previous convolutional block is considered only to be the one from which $l$ has incoming links. The distinction is made for simplicity in computation, as the statistics of the previous layer  is only required in case (b) of our approach (Section ~\ref{sec_approach}), for deciding whether any operation should be applied to the current block or not.  }
\label{fig_nets_overview}
\end{figure} 
 
\textbf{Transfer learning with deep nets: } With the availability of large scale datasets such as ImageNet \cite{deng2009imagenet} and MIT Places \cite{zhou2014learning}, researchers have resorted to transfer learning techniques \cite{torrey2009transfer}  for efficient training of relatively smaller related datasets \cite{oquab2014learning, zhou2014hybrid}. During transfer learning, the parameters of the CNN trained with base dataset are duplicated, and some additional layers are attached at the deep end of the CNN which are trained exclusively on the new dataset. In the process, the parameters copied from the net trained on the base dataset might or might not be allowed for slight perturbation. However, none of the transfer learning techniques attempts to refine the CNN architecture effecting an increase in original accuracy and a reduction in model size simultaneously.  While transfer learning can be effective when the base dataset has a similar distribution as the target dataset, it might be a deterrent otherwise \cite{Shankar_2015_CVPR}. We emphasize that our approach can be applied to any pre-trained CNN, irrespective of whether the training has been done through transfer learning or from scratch. 

\textbf{Low-rank and sparsification methods for CNNs: } Irrespective of whether a CNN has been hand-designed for a specific dataset or not, all the famous CNN architectures exhibit enormous redundancy in parameter space \cite{liu2015sparse}. Researchers have recently exploited this fact to speed up the inference speeds by estimating a highly reduced set of parameters which is sufficient to produce nearly the same accuracy as the original CNN does with the full  set of parameters. While some works like \cite{sainath2013low,jaderberg2014speeding, xue2013restructuring} have resorted to low-rank factorization of weight matrices between two given layers, others have used sparsification methods for the same \cite{graham2014spatially}. Recently, \cite{liu2015sparse} has combined the low-rank and sparsification methods to produce a highly sparse CNN with a slight decrease in the original accuracy. The work of \cite{denil2013predicting} can be considered as a pseudo-reduction method for the parameter space of a CNN. It does not sparsify the network, but presents an approach to estimate almost $95\%$ of parameters from only the rest $5\%$. Thus, they do not claim that most parameters are not necessary, but that most parameters can be estimated by a relatively small set. 


It is worthwhile to mention that our approach falls into a different solution paradigm, that can complement various methods developed for deep learning for distinct purposes. All the related works discussed above and some other works that tend to enhance the accuracy with deep learning such as \cite{Shankar_2015_CVPR} and deep boosting methods \cite{shalev2014selfieboost, cortes2014deep, peng2014deep}, assume a fixed architectural model of the CNN. Our approach instead modifies the architecture of the original CNN to potentially enhance the accuracy while possibly reducing the model size. Thus, all the techniques applied to a fixed architecture can be applied to the architecture refined by our method, for a plausibly better performance as per the chosen metric. Also, due to the novel operations that we consider for CNN architectural refinement, our method can complement the various other methods developed for a similar purpose. 

\section{Approach} \label{sec_approach}
Let the dataset contain $M$ classes. Let the CNN architecture have $L$ convolutional layers. At a given convolutional layer $l \in \{1,\dots,L\}$, let there be $h_l$ number of hidden units (nodes). Then for a given input image $i$, one can generate an $h_l$ dimensional feature vector $\boldsymbol{f_l^i}$ at convolutional layer $l$, by taking average spatial responses at each hidden unit of the layer \cite{zhou2014object}. Using this, one can find a mean feature vector of dimension $h_l$ for every class $m \in \{1, \dots, M\}$ at every convolutional layer $l$ by taking the average of $\boldsymbol{f_l^i} ~~\forall ~~i \in \boldsymbol{a_m}$, where the set $ \boldsymbol{a_m}$ contains images annotated with class label $m$.  Let this average feature vector for class $m$  be denoted by $\boldsymbol{g_l^m}$. 

\textbf{Finding the inter-class separation:}
For a given dataset and a base CNN architecture, we first train the CNN on the dataset using a given loss function (such as softmax loss, sigmoid cross entropy loss, etc.  \cite{jia2014caffe}). From this pre-trained CNN, we compute  $\boldsymbol{g_l^m}; l \in \{1,\dots,L\}, m \in \{1, \dots, M\}$. Using $\boldsymbol{g_l^m}$, inter-class correlation matrices $\boldsymbol{C_l}$ of sizes $M \times M$ are found out for every convolutional layer $l$, where a value at the index-pair $(m,\hat{m}) ; m, \hat{m} \in \{1, \dots, M\}$  in $\boldsymbol{C_l}$ indicates the correlation between $\boldsymbol{g_l^m}$ and $\boldsymbol{g_l^{\hat{m}}}$. Note that the correlation between two feature vectors of the same length can vary between -1 and 1, inclusive. Examples of  $\boldsymbol{C_l}$ can be seen in Fig ~\ref{fig_dataset_characteristics}. All $\boldsymbol{C_l}$ are symmetric, since correlation is non-causal. A lesser correlation between classes implies better separation, and vice-versa. 

\textbf{Measuring separation enhancement and deterioration capacity of a layer:} The correlation matrices give an indication of the separation between classes for a given convolutional layer. Comparing $\boldsymbol{C_l}$ and $\boldsymbol{C_{l+1}}$, one can know for which class pairs, the separation increased (correlation decreased) in layer $l+1$, and for which ones, the separation deteriorated (correlation increased) in $l+1$.  Similar statistics for layer $l$ can be computed by comparing $\boldsymbol{C_{l-1}}$ and $\boldsymbol{C_l}$.  For a convolutional layer $l$, let the number of class pairs where the separation increased in comparison to layer $l-1$ be $n_+^l$, and where the separation decreased be $n_-^l$. Let $n_T = M^2$ denote the total number of class pairs. Note that both stretch and split operations can be simultaneously applied to each layer. $n_+^l$ contributes to the \textit{stretching / widening} of the layer $l$, while $n_-^l$ contributes to the \textit{symmetric splitting} of its inputs. 

In the domain of decision tree training \cite{criminisi2012decision}, information gain is used to quantify the value a node adds to the classification task. However, in the context of our work, this measure would not enable us to estimate both the split and the stretch factors for the same layer. Thus, we resort to the number of class pairs where the separation increases / decreases to measure the separation enhancement and deterioration capacity of a layer respectively. As we will discuss in the next subsection and Section ~\ref{sec_results}, both the stretch and split operations applied to the same layer helps us to optimally reduce the model size and increase accuracy.   

\begin{figure*}[!htb]
\begin{center}
   \includegraphics[width=0.95\linewidth, height=0.14\textheight]{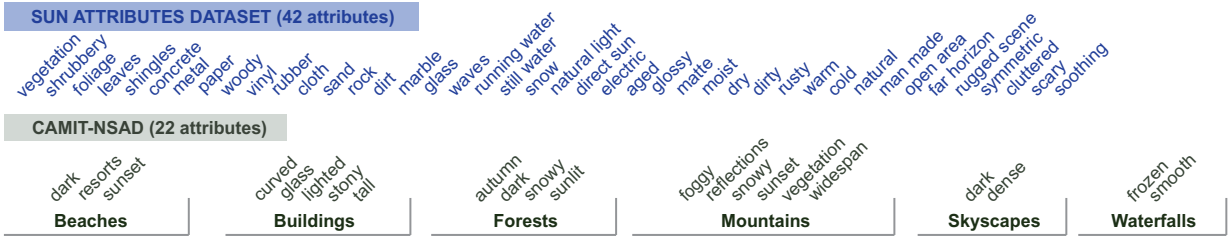}
\end{center}
   \caption{\fontsize{8.5pt}{10pt}\selectfont  \textbf{Datasets and Classes:  } The figure shows the classes present in the SUN Attributes Dataset (SAD) as considered in \cite{Shankar_2015_CVPR} and CAMIT-NSAD dataset \cite{Shankar_2015_CVPR}. While classes in SAD are purely attributes, classes in CAMIT-NSAD are attribute-noun pairs.}
\label{fig_attribute_choices}
\end{figure*}

\textbf{Estimating stretch and split factors:} By the definition of $n_+^l$, for a layer $l$, we define the average class separation enhancement capability of the subsequent layers by the following expression: 
\begin{equation}  \label{equ_0}
 \xi(l) = \sum_{i = {l}+1}^{L-1} (n_+^i ~/~ n_T) / (L - {l} - 1)  
\end{equation}
Note that we omit the last layer $L$ in the above expression. This is discussed at the end of this subsection. 

\textit{For each $l$, there can be two cases, (a) $n_+^l < n_-^l$, (b) $n_+^l \geq n_-^l$.} \textbf{Case (a)} implies that the number of class pairs for which separation decreased were more than for which separation increased. This is not a desired scenario, since with subsequent layers, we would want to gradually increase the separation between classes. Thus, in case (a), we do a symmetric split between $l-1$ and $l$, i.e. the connections incoming to $l$ undergo a split. This is done under the hypothesis that split should minimize the hindering linkages and thus cause a lesser deterioration in the separation of class pairs in layer $l$. The amount of split is decided by $n_-^l$ and the average separation enhancement potential of the subsequent layers $\xi(l)$. For example, if the subsequent layers greatly increase the separation between classes, a lesser split should suffice since we do not need to improve the separation potential between layers $l$ and $l+1$ by a major extent. Doing a high amount of split in this case may be counter-productive, since the efficient subsequent layers might not then get sufficiently informative features to proceed with. Based on this hypothesis, we arrive at the following equation indicating the split factor $r_s^l$ for convolutional layer $l$ under case(a):
\begin{equation} \label{eqn_1}
r_s^l =2^{ \psi \left(\frac{n_-^l}{n_T} ~~  \xi(l)  \right) }
\end{equation} 
where $\psi(x) = \lfloor   x / \lambda \rfloor$. $\lambda$ is a parameter that controls the amount of reduction in model size. Note that the expression is raised to the power of $2$ in \eqref{eqn_1}, meaning that we do splits in multiples of $2$. This is done to make the implementation coherent for Caffe's \cite{jia2014caffe} group parameter. The group parameter in Caffe is similar to the symmetric split operation  considered here (Fig ~\ref{fig_refinement_operations}). Although \textit{group} parameter can be any integer, it should exactly divide the number of nodes being split. Since,  the number of nodes in architecture layers are typically multiples of 2, we raise the expression to the power of 2 in \eqref{eqn_1}. For case (a), no stretching is performed, since that might lead to more redundancy. 

\textbf{For case (b), } the number of class pairs experiencing increased separation are greater than those undergoing deterioration. We aim to stretch the layer as well as split its inputs in such a scenario. The stretch factor is based on $n_+^l$ and the average separation enhancement capability of subsequent layers $\xi(l)$. If $\xi(l)$ is significant, stretching in $l$ is done to a lesser extent indicating that $l$ needs to help but only to a limited extent to avoid overfitting; and vice-versa. We thus arrive at the following equation indicating the stretch factor $r_e^l$  for layer $l$ in case (b):
\begin{equation} \label{eqn_2}
r_e^l = 1 + \phi \left(\frac{n_+^l}{n_T} ~~  \xi(l)  \right)
\end{equation} 
where $\phi(x) = \lfloor   x / \lambda \rfloor ~\lambda$ is a function that depends on $\lambda$. We add $1$ in \eqref{eqn_2}, since a stretch factor of say 1.25 indicates that the number of nodes in the respective layer be increased by a quarter. Note that $\phi(x) = \lambda \psi(x)$. This indicates that for say $\lambda = 0.25$, a split factor of $2$ might be roughly equivalent to a stretch factor of $1.25$ for enhancing the class separation. This is an empirical choice, which helps us to optimally increase the accuracy and reduce the model size. We will delineate the importance of $\lambda$ in the next subsection. 

In case (b), due to $n_-^l$, there is also some redundancy in the connections between $l$ and $l-1$. Thus the inputs of layer $l$ also need to be split. The split factor in this case is again decided by \eqref{eqn_1}. The operation of splitting along with stretching helps to reduce the model size while also potentially enhancing the accuracy.



In our approach, we do not consider the refinement of fully connected layers, but only refine the convolutional layers. This is motivated by the fact that in CNNs, convolutional layers are mostly present in high numbers, with fully connected layers being lesser in number.  For instance, GoogleNet has only one fully connected layer at the end after 21 convolutional layers. However, since fully connected layers can contain a significant amount of parameters in comparison to convolutional layers (like in AlexNet), considering fully connected layers for architectural refinement can be worth exploring. 

Since for a layer, our method considers the change in class separation compared to the previous layer, no stretching or splitting is done for the first convolutional layer since it is preceded by the input layer. Also, we notice that the final convolutional layer in general, enhances the separation for most classes in comparison to the penultimate convolutional layer. Thus, stretching it mostly amounts to overfitting, and so, we exclude the last convolutional layer from all our analysis. By a similar argument, the last inception unit is omitted from our analysis in GoogleNet. 

Once the stretch / split factors are found using a pre-trained architecture, the refined architecture is trained from scratch. Thus, we do not share any weights between the original and the refined architecture. The weight initialization in all cases is done according to \cite{he2015delving}. 

\textbf{On choice of $\lambda$ and upper bound:} The parameter $\lambda$ controls the amount of reduction in model size. It is meant to be empirically chosen and the functions $\psi(.), \phi(.) $ have been formulated so that they satisfy our hypotheses mentioned in the previous subsection while taking into account the possible effect of $\lambda$. If $\lambda$ increases, the stretch factors decrease and the splits are smaller (see $\phi(.)$ and $\psi(.)$). If $\lambda$ decreases, the split factors can be very high along with decent values for stretch factors. However, due to the difference in $\phi(.)$ and $\psi(.)$, the increase in the stretch factor will be limited as compared to the increase in the split factor. This is desired since we do not wish to increase the model size by vast amounts of stretching operations, which may also lead to overfitting. \textit{Hence, with increasing $\lambda$, the model size tends to increase, and vice-versa.} For all our experiments, we set an empirically chosen $\lambda = 0.25$. 

A natural question to now ask is that what range of values of $\lambda$ should one try? The lower bound may be empirically chosen based on the maximum split factor that one wishes to support. However, $\lambda$ can be upper-bounded by a value $\lambda_o$ above which no split and stretch factor can change. From the definitions of $\psi(.)$ and $\phi(.)$, $\lambda_o$ can be easily given as follows:
\begin{equation}  \label{eqn_4}
\lambda_o = max \left[ \left(\frac{n_+^{l'}}{n_T} ~~ \xi(l')    \right), \left(\frac{n_-^{l'}}{n_T} ~~   \xi(l') \right), \left(\frac{n_-^{l}}{n_T} ~~ \xi(l)    \right)  \right]
\end{equation}
where $~ n_+^{l'} \geq n_-^{l'} ~~\forall ~ l' ; n_+^{l} < n_-^{l} ~~ \forall ~l $.

\textbf{Refining with GoogleNet: }Note that GoogleNet contains various inception modules (Fig~\ref{fig_nets_overview}), each module having two layers with multiple convolutional blocks. While describing our refinement algorithm, wherever we have mentioned the term \textit{convolutional layer}, in context of GoogleNet, it should be considered as a convolutional block. Please see Fig ~\ref{fig_nets_overview} for a better understanding of how do we decide subsequent and previous layers in GoogleNet for refining. Also see Fig ~\ref{fig_arch_desc} for the stretch and split factors obtained after architectural refinement of GoogleNet. 

\textbf{Continuing architectural refinement:} Our method refines the architecture of a pre-trained CNN. One can also apply such a procedure to an architecture that has been already refined by our approach. One can stop once no significant difference in the accuracy or model size is noticed for some choices of $\lambda$ \footnote{Although our approach does not induce a concrete optimization objective from the correlation analysis, we believe that it is a step towards solving the deep architecture learning problem, and furthers the related works towards more principled directions. The intuition behind our method was established from various experiments, done on diverse datasets with a variety of shallow and deep CNNs.}.

\section{Results and Discussion} \label{sec_results}


\begin{figure*}[!htb]
\begin{center}
   \includegraphics[width=0.95\linewidth, height=0.18\textheight]{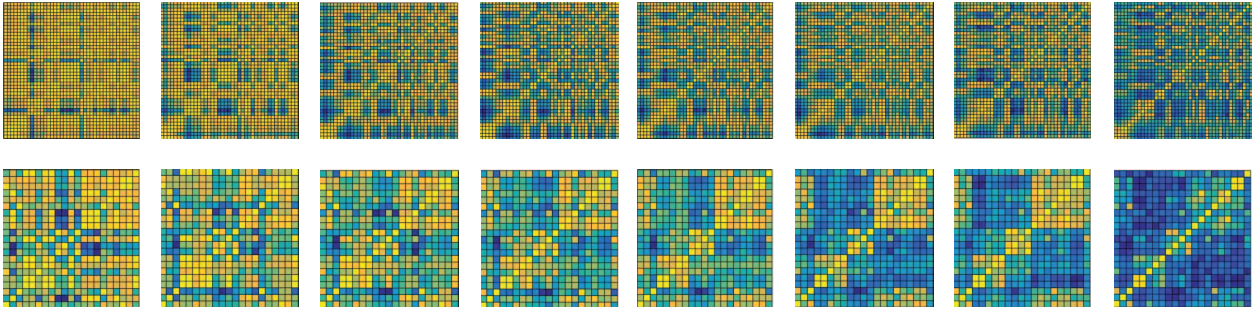}
\end{center}
   \caption{\fontsize{8.5pt}{10pt}\selectfont  \textbf{Correlation Matrices for $8$ Convolutional Layers of VGG-11 trained on SAD and CAMIT-NSAD :  } Traversed row-wise, correlation matrices $\boldsymbol{C_l}: l \in \{1, \dots,  8\}$ are shown.  Dark Blue color indicates minimum correlation between classes, while a bright yellow color indicates maximum correlation. Thus, all diagonals are bright yellow, since each class is maximally correlated with itself. For each matrix, the attribute classes are ordered as in Fig ~\ref{fig_attribute_choices} seen left to right. Note that more correlation implies lesser separation and vice-versa. \textbf{\textit{Top Row  (SAD) } :}   The lower layers can separate the classes better as compared to deeper layers. \textbf{\textit{Bottom Row (CAMIT-NSAD)}:}  The classes are separated lesser in lower layers and more prominently in deeper layers. This is mainly because classes in SAD are purely attributes, while classes in CAMIT-NSAD are attribute-noun pairs. Due to this distinction, the two datasets have nearly contrasting characteristics which pose a challenge to the architectural refinement problem. For instance, \textit{snow} class in SAD can be separated from \textit{dirt} class mostly by the distinction of white and brown colors; while in CAMIT-NSAD, the class of \textit{snowy forests} cannot be separated from \textit{snowy mountains} just by noticing the color difference, since both forests and mountains are snowy. Infact, in this case, the separation is most likely to appear in deeper layers where the distinction is also made between forests and mountains. The above explanation is made under the widely accepted notion that a CNN learns low-level type features (edges, color patterns, etc.) in lower layers, and more class-specific features in deeper layers \cite{yosinski2014transferable}.  Also note that some classes in the datasets have a natural correlation, e.g. classes of \textit{vegetation}, \textit{shrubbery}, \textit{foliage} and \textit{leaves} in SAD are well correlated, since the presence of leaves is very likely where some vegetation occurs. As a result, separation between these classes may always be low as compared to separation between the classes of \textit{vegetation} and \textit{running water}. A similar analysis can be made for CAMIT-NSAD. \textit{Figure is best viewed in color.}}
\label{fig_dataset_characteristics}
\end{figure*}

\textbf{Datasets:} We evaluate our approach on SUN Attributes Dataset (SAD) \cite{patterson2012sun} and  Cambridge-MIT \textbf{N}atural \textbf{S}cenes \textbf{A}ttributes \textbf{D}ataset (CAMIT-NSAD) \cite{Shankar_2015_CVPR}. Both the datasets have classes of natural scenes attributes, whose listing can be found in Fig ~\ref{fig_attribute_choices}. 

The full version of SAD \cite{patterson2012sun} has 102 classes. However, following \cite{Shankar_2015_CVPR}, we discard the classes in the full version of SAD which lie under the paradigm of recognizing activities, poses and colors; and only consider the 42 visual attribute classes, so that the dataset  is reasonably homogeneous for our problem. SAD with 42 attributes has 22,084 images for training, 3056 images for validation and 5618 images for testing. Each image in the training set is annotated with only one class label. Each test image has a binary label for each of the 42 attributes, indicating the absence / presence of the respective attribute. In all, the test set contains 53,096 positive labels.  

CAMIT-NSAD \cite{Shankar_2015_CVPR} is a natural scenes attributes dataset containing classes as attribute noun pairs, instead of just attributes. CAMIT-NSAD has 22 attributes and contains 46,008 training images, with at least 500 images for each attribute-noun pair. The validation set and the test set contain  2104 and 2967 images respectively. While each training image is annotated with only one class label, the test images contain binary labels for each of the 22 attributes. In all, the test set contains 8517 positive labels. All images in SAD and CAMIT-NSAD are $256 \times 256$ RGB. 

It can be seen that classes in SAD are pure attributes, while that in CAMIT-NSAD are noun-attribute pairs. Due to this distinction, the two datasets have different characteristics which make them challenging for the problem of architectural refinement. Please see Fig ~\ref{fig_dataset_characteristics} for a better understanding of this distinction.



\begin{figure*}[!htb]
\begin{center}
   \includegraphics[width=0.95\linewidth, height=0.16\textheight]{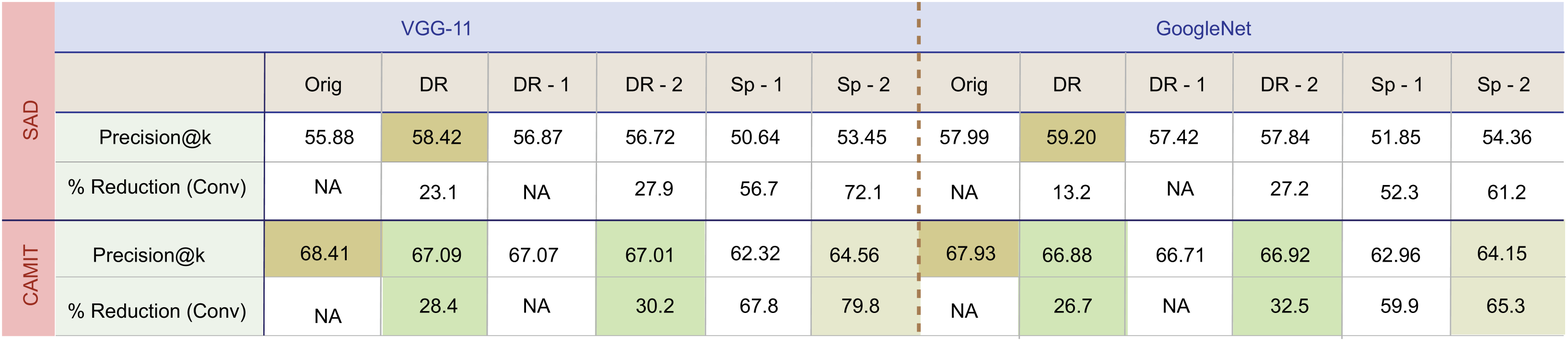}
\end{center}
   \caption{\fontsize{8.5pt}{10pt}\selectfont  \textbf{Results and Comparisons : } The figure shows results on SAD and CAMIT-NSAD with VGG-11 and GoogleNet using our approach and various other baselines. \textit{Orig} = Original architecture, \textit{DR} = Deep Refined Architecture (our approach), \textit{DR-1} = Deep Refined Architecture with only the Stretch Operation,  \textit{DR-2} = Deep Refined Architecture with only the Symmetric Split Operation,  \textit{Sp-1} = L1 Sparsified network, \textit{Sp-2} - Sparsified network with \cite{liu2015sparse}. We use \textit{precision@k} as our performance metric, and report that here as a percentage. For SAD, k = 21, while for CAMIT-NSAD, k = 7.  Please refer text for details on this. We report the reduction percentage in the parameters of the convolutional layers in comparison to the original architecture. Thus, reporting reduction in model size for $Orig$ is not applicable. Also, since \textit{DR-1} only does a stretching operation over the original architecture, it is bound to increase the model size, and thus $\%$ reductions in the model size are not applicable here as well. Note that \textit{DR} performs significantly well for SAD giving a decent reduction in model size with impressive increase in precision. For CAMIT-NSAD, \textit{DR} does not improve the precision of the original architecture. The results here are reported for $\lambda = 0.25$. For CAMIT-NSAD, we also did experiments for higher values of $\lambda$ but we did not see any increase in precision; rather with increased $\lambda$, we got lesser reduction in model size as expected (Section ~\ref{sec_approach}) . Nevertheless, for CAMIT-NSAD, \textit{DR} and \textit{DR-2} could produce architectures with a reduced model size producing precision better than the state-of-the-art sparsification techniques of \textit{Sp-1} and \textit{Sp-2}.}
\label{fig_res_main}
\end{figure*}

\begin{figure*}[!htb]
\begin{center}
   \includegraphics[width=0.98\linewidth, height=0.21\textheight]{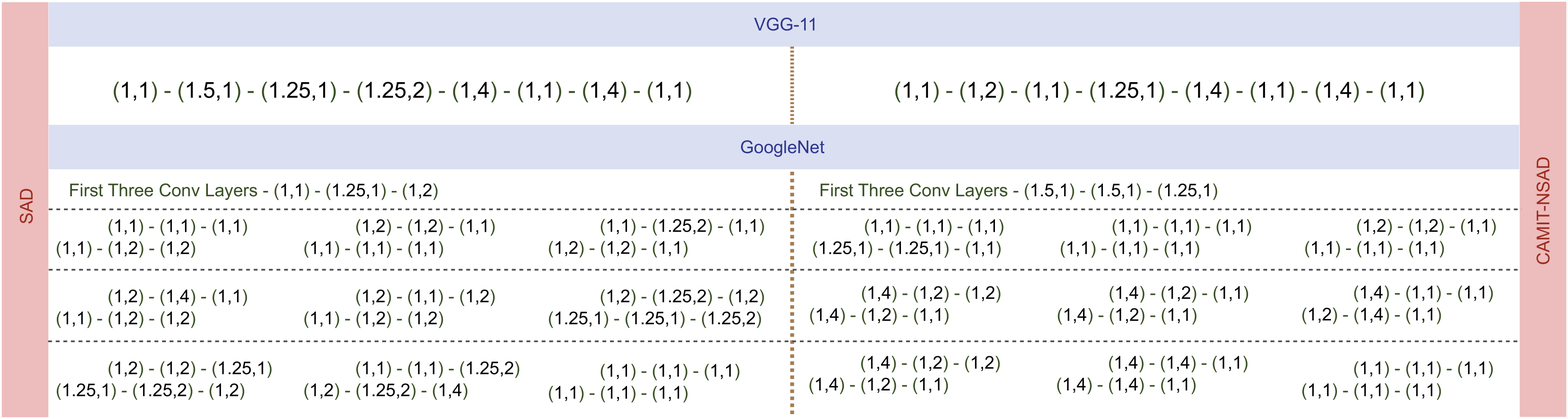}
\end{center}
   \caption{\fontsize{8.5pt}{10pt}\selectfont  \textbf{Refined Architectures obtained with our approach: }  Left column shows the refined architectures for SAD, and the right column for CAMIT-NSAD. The corresponding precision results are reported in Fig ~\ref{fig_res_main} under the column \textit{DR}. Each tuple $(a,b)$ indicates that $a$ is the stretch factor for the convolutional layer/block, while $b$ is the split factor for the input of that convolutional layer / block. Entry of $(1,1)$ implies no stretch and splitting should be done. In Fig ~\ref{fig_res_main}, for \textit{DR-1} , every value of $b$ is made 1, while for  \textit{DR-2} , every value of $a$ is made 1.  VGG-11 contains 8 convolutional layers, for which the factors are shown. In Googlenet, the factors for first three convolutional layers are shown in the first row (under GoogleNet). After that , each row under GoogleNet contains the factors of the convolutional blocks in the inception unit of Fig ~\ref{fig_nets_overview}. Traversed row-wise, inception units correspond to the ordering 3a, 3b, 4a, 4b, 4c, 4d, 4e, 5a, 5b of GoogleNet architecture \cite{szegedy2014going}. Note that since we do not consider the last convolutional layer (that is connected to the fully connected layer) in our analysis, all factors for that are 1 in VGG-11. A similar argument exists for the last inception unit in GoogleNet. }
\label{fig_arch_desc}
\end{figure*}

Notice that classes in CAMIT-NSAD can be finally separated to a greater extent as compared to classes in SAD. This is because almost each class in SAD has a variety of outdoor and indoor scenes, since an attribute can exist for both. For instance,  both an outdoor and indoor scene can be \textit{glossy} as well as can have  \textit{direct sunlight}. However, with noun-attribute pairing as in CAMIT-NSAD, the classes are more specifically defined, and thus significant separation between a greater number of class pairs is achieved at the end. 

\textbf{Choice of Datasets: } The choice of the datasets used for evaluation needs a special mention. We chose attribute datasets, since 
given the type of labels here, it is difficult to establish where should the model parameters be reduced / increased. This we found was in contrast to \textit{object} recognition datasets such as ImageNe, where we observed that refining an architecture by symmetric splitting in the first few layers could increase accuracy. However, we thought this to be very intuitive, since \textit{objects} are generally encoded in deeper layers, and thus, one would expect to reduce parameters in the top layers. We thus evaluate our procedure with the types of datasets, where one cannot easily decide which network layers contribute to the class labels. 

\textbf{Architectures for Refinement: } We choose GoogleNet \cite{szegedy2014going} and VGG-11 \cite{simonyan2014very} as the base CNN architectures, which we intend to alter using our approach. Since GoogleNet and VGG-11 are quite contrasting in their construction, they together pose a considerable challenge to our architectural refinement algorithm. While VGG-11 (which can be loosely considered as a deeper form of AlexNet \cite{krizhevsky2012imagenet}) has $8$ convolutional layers, and 3 fully connected layers, GoogleNet is a 22-layer deep net having $9$ inception units  after three convolutional layers,  and a fully connected layer before the final output. Each inception unit has $6$ convolutional blocks arranged in $2$ layers. We refer the reader to \cite{szegedy2014going} and \cite{simonyan2014very} for complete details of GoogleNet and VGG-11 respectively. An instance of inception module in GoogleNet is shown in Fig ~\ref{fig_nets_overview}. 

\textbf{Baselines:}
We consider the following baselines to compare with our proposed approach. \textit{\textbf{(a) Our approach with only stretching and no splitting}} - We consider refinement with only the stretching operation and no splitting operation, i.e. the CNN architecture is refined by only stretching some layers, but no symmetric splitting between layers is done. This proves the importance of stretching operation for architectural refinement. \textit{\textbf{(b) Our approach with only splitting and no stretching}} - Evaluating by only considering the symmetrical split operation and no stretch operation provides evidence to the utility of splitting. \textit{\textbf{(c) L1 Sparsification}} - We consider the L1 sparsification of a CNN as one of the important baselines. Here, the weights (parameters) of a CNN are regularized with an L1 norm, and the regularization term is added to the loss function. Due to the L1 norm, this results in a sparse CNN, i.e. a CNN with a reduced model size. Following \cite{liu2015sparse}, all the parameters with values less than or equal to \textit{1e-4} are made zero both during training and testing. This not only ensures maximal sparsity, but also stabilizes the training procedure resulting in better convergence. \textit{\textbf{(d) Sparsification with the method of \cite{liu2015sparse} }-  } The method of \cite{liu2015sparse} combines the low-rank decomposition \cite{jaderberg2014speeding} and L1 sparsification techniques for better sparsity. However, they mention that the critical step in achieving comparable accuracy with high amount of sparsity, is minimizing the loss function along with L1 and L2 regularization terms upon the weights of the CNN. Low-rank decomposition can increase sparsity with a further decrease in accuracy. Since, in this work, we are interested in an optimal trade-off between accuracy and model size, we evaluate the method of \cite{liu2015sparse} without the low-rank decomposition. This ensures that we obtain maximum possible accuracy with \cite{liu2015sparse} at the expense of some reduced sparsity. \textit{\textbf{(e) Original architecture:}} We consider the original architecture without any architectural refinement and sparsification techniques applied. The amount of reduction achieved in the model size with our approach and other baselines along with the recognition performance is compared with this baseline.

Note that for all the above mentioned baselines, the CNN is first trained on the respective dataset with the standard minimization of softmax loss function \cite{jia2014caffe}, after which a second training step is done. For baselines (a) and (b), the re-training step is performed on the refined architecture as described in Section ~\ref{sec_approach}; while for baselines (c) and (d), retraining is done as a fine-tuning step, where the learning rate of the output layer is increased to 5 times the learning rate of all  other layers. 

\textbf{Other plausible baselines:} We also tried randomly splitting and stretching throughout the network as a plausible baseline.  Here although in some cases, we could reduce the model size by almost similar amounts as our proposed approach, significantly higher accuracy was consistently achieved using our method. 

\textbf{Training:}
For all datasets and CNN architectures, the networks are trained using the Caffe library \cite{jia2014caffe}. The pre-training step is always performed with the standard softmax loss function \cite{jia2014caffe}. For all the pre-training, refinement and baseline cases, batch size of 32 samples is considered. An \textit{adaptive step policy} is followed during training, i.e. if the change in validation accuracy over a range of 5 consecutive epochs is less than $0.5$, the learning rate is reduced by a factor of 10. For SAD, we start with an initial learning rate of $0.01$ for both GoogleNet and VGG-11, while for CAMIT-NSAD, a starting learning rate of $0.001$ suffices for both the architectures. In all cases, we train for 100 epochs. 

\textbf{Testing:} Given a trained CNN, we need to predict multiple labels for each test image in SAD and CAMIT-NSAD. We use \textit{precision@k} as our performance metric. The metric is normally chosen when one needs to predict top-k labels for a test image. Since, our ground-truth annotations contain only binary labels for each class, for a given test image, we cannot sort the labels according to their degree of presence. We thus decide $k$ for each dataset as the maximum number of positive labels present for any image in the test set. For SAD, $k$ is 21, while for CAMIT-NSAD, $k$ is 7. Thus, given a test image, we predict the output probabilities of each class using the trained net, and sort these probabilities in the descending order to produce a vector $\boldsymbol{T}$. If that test image has say 5 positive labels in ground-truth annotations, we expect the first 5 entries of $\boldsymbol{T}$ to correspond to the same labels for a $100\%$ precision. We thus compute the true positives and false positives over the entire test set and report the final precision.  This is in line with the test procedure followed by \cite{Shankar_2015_CVPR}. 

\textbf{Discussion of results:}
Fig ~\ref{fig_res_main} shows the precision and model size reduction results obtained with our approach and the baselines, for both the datasets and both the architectures. For understanding intrinsic details of the refined architectures, please refer to Fig ~\ref{fig_arch_desc} and Fig ~\ref{fig_nets_overview}. It is clear that for SAD, our approach for both VGG-11 and GoogleNet, offers an increase in original precision while giving a reasonable reduction in model size. The reduction in the number of parameters in convolutional layers holds more importance here, since our method was only applied to the convolutional layers. It is interesting to note from the results on SAD, that the predicted combination of stretch and split is more optimal as compared to only having the split or the stretch operation. This also shows that stretching alone is not always bound to enhance the precision, since it may lead to  overfitting. In all cases, the sparsification baselines fall  behind the precision obtained with our approach, although they produce more sparsity. 

The results on CAMIT-NSAD present a different scenario. Note that our approach is not able to enhance the precision in this case, but decreases the precision by a small amount, while giving decent reduction in model size. However, the precision obtained with a reduced model size by using our approach  is still greater than the one obtained by other baseline sparsification methods, though at the expense of lesser sparsity. The inability to increase precision in this case can be attributed to the fact that our approach is greedy, i.e. it estimates the stretch and split factors for every layer, and not jointly for all layers. This affects CAMIT-NSAD since the classes are attribute-noun pairs, and attribute-specific information and noun-specific information are encoded at different layers, which need to be considered together for refinement. 

Note that a single metric jointly quantifying both the accuracy increase and the model size reduction is difficult to formulate. 
In cases where we increase the accuracy as well as decrease the model size (SAD), we offer a win-win situation. However, in cases where we decrease the model size but cannot increase the accuracy (CAMIT-NSAD), we believe that the model choice depends on user's requirements, and  our method provides an additional and plausibly a useful alternative  for the user, and can also complement the other approaches. One can obtain an architecture using our approach, and then apply a sparsification technique like \cite{liu2015sparse} in case the user's application demands maximum sparsity, and not that good a precision. 

\section{Conclusion} \label{sec_conclusions}
We have introduced a novel strategy that alters the architecture of a given CNN for a specified dataset  for effecting a possible increase in original accuracy and reduction of parameters. Evaluation on two challenging datasets shows its utility over relevant baselines.





{\small
\bibliographystyle{ieee}
\bibliography{mainBib}
}

\end{document}